\documentclass[journal]{IEEEtran}
\usepackage{comment}

 \usepackage{color}
\usepackage{cite}

\ifCLASSINFOpdf
   \usepackage[pdftex]{graphicx}
\else
\fi
\usepackage{amsmath,amssymb}
\ifCLASSOPTIONcompsoc
  \usepackage[caption=false,font=normalsize,labelfont=sf,textfont=sf]{subfig}
\else
  \usepackage[caption=false,font=footnotesize]{subfig}
\fi
\begin{document}
%
\title{Continual Learning Using Bayesian Neural Networks}
%
%
%

\author{Honglin Li,
        Payam Barnaghi,~\IEEEmembership{Senior Member IEEE},
        Shirin Enshaeifar,~\IEEEmembership{Member IEEE},
        Frieder Ganz 
\thanks{H. Li, P. Barnaghi and S. Enshaeifar are with the Centre for Vision, Speesh and Signal Processing (CVSSP) at the University of Surrey and also with Care Research and Technology Centre at the UK Dementia Research Institute (UK DRI). P. Barnaghi is also with the Department of Brain Sciences at Imperial College London. email: \{h.li,s.enshaeifar\}@surrey.ac.uk, {p.barnaghi}@imperial.ac.uk}
\thanks{F. Ganz is with the Adobe, Germany. email: {ganz}@adobe.com}}
%
%

\markboth{Journal of \LaTeX\ Class Files,~Vol.~14, No.~8, August~2015}%
{Shell \MakeLowercase{\textit{et al.}}: Bare Demo of IEEEtran.cls for IEEE Journals}
%



\maketitle

\begin{abstract}
Continual learning models allow them to learn and adapt to new changes and tasks over time. However, in continual and sequential learning scenarios in which the models are trained using different data with various distributions, neural networks tend to forget the previously learned knowledge. This phenomenon is often referred to as catastrophic forgetting. The catastrophic forgetting is an inevitable problem in continual learning models for dynamic environments. To address this issue, we propose a method, called Continual Bayesian Learning Networks (CBLN), which enables the networks to allocate additional resources to adapt to new tasks without forgetting the previously learned tasks. Using a Bayesian Neural Network, CBLN maintains a mixture of Gaussian posterior distributions that are associated with different tasks. The proposed method tries to optimise the number of resources that are needed to learn each task and avoids an exponential increase in the number of resources that are involved in learning multiple tasks. The proposed method does not need to access the past training data and can choose suitable weights to classify the data points during the test time automatically based on an uncertainty criterion. We have evaluated our method on the MNIST and UCR time-series datasets. The evaluation results show that our method can address the catastrophic forgetting problem at a promising rate compared to the state-of-the-art models.
\end{abstract}


%
\IEEEpeerreviewmaketitle

\section{Introduction}
%
%
%
%
\IEEEPARstart{D}{eep} learning models provide an effective end-to-end learning approach in a variety of fields. One common solution in deep neural networks to solve a complex task such as ImageNet Large Scale Visual Recognition Challenge (ILSVRC) \cite{deng2009imagenet} is to increase the depth of the network \cite{he2016deep}. However, as the depth increases, it becomes harder for the training model to converge. On the other hand, a shallower network is not able to solve a complex classification task at once, but it may be able to find a solution for a smaller set of classes and converges much faster. If a model can continually learn several tasks, then it can solve a complex task by dividing it into several simple tasks. In online learning scenarios \cite{shalev2011online}, the model repeatedly receives new data, and the training data is not complete at any given time. If we re-train the entire model whenever there are new instances, it would be very inefficient, and we have to store the trained samples \cite{de2019continual}. The key challenge in such continual learning scenarios in changing environments is how to incrementally and continually learn new tasks without forgetting the previous or creating highly complex models that may require accessing the entire training data. 

Most of the common deep learning models are not capable of adapting to different tasks without forgetting what they have learned in the past. These models are often trained via back-propagation where the weights are updated based on a global error function. Updating and altering tasks of an already learned model leads to the loss of the previously learned knowledge as the network is not able to maintain the important weights for various distributions. If hard constraints are applied to the model to prevent the forgetting, it can retain the previously learned knowledge. However, the model will not be able to acquire new information efficiently. This scenario is referred to as the stability-plasticity dilemma \cite{mermillod2013stability,ditzler2015learning}. The attempt to sequentially or continuously learn and adapt to various distributions will eventually result in a model collapse. This phenomenon is referred as catastrophic forgetting or interference \cite{mccloskey1989catastrophic,goodfellow2013empirical}. The catastrophic forgetting problem makes the model inflexible. Furthermore, the need for a complete set of training samples during the learning process is very different from the normal biological systems which can incrementally learn and acquire new knowledge without forgetting what is learned in the past \cite{mcclelland1995there,barnett2002and}. 

\begin{figure}
    \centering
       \includegraphics[width=\linewidth,height=4cm]{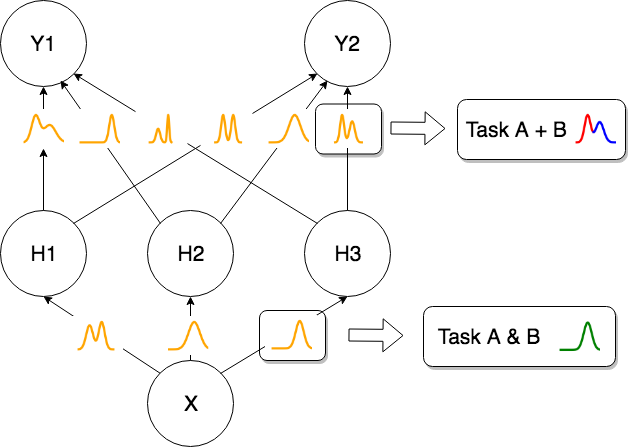}
    \caption{The network architectures. In Continual Bayesian Learning networks, the weights are Gaussian mixture distributions with an arbitrary number of components. e.g. As shown above, two Bayesian Neural Networks which have learned Task A and B respectively. Each of them contains multiple Gaussian distributions. The CBLN merges these Gaussian distributions into one Gaussian mixture distribution. The number of components in the mixture distribution in this example can be 2, which means task A and task B have different weight distributions (shown in red and blue), or it could be 1, which means tasks A and B have the same weight distributions (shown in green).}
    \label{fig:network_archi}
\end{figure}

\textbf{Terminology:} In this paper, the term \textit{task} refers to the overall function of a model; e.g. classification, clustering and outlier detection.  A task has an input distribution and an output distribution. A \textit{dataset} is used to train and evaluate a model for a task. A dataset follows a certain distribution. The \textit{distribution} of a dataset that is used to train a specific task can change over time. We can train a model with different tasks. Each task can be trained on its own individual dataset. In other words, each task can be trained based on different input and output distribution. 

To address the catastrophic forgetting problem, there are mainly three approaches \cite{parisi2019continual}:

\textbf{Regularisation Approaches:} Regularisation based approaches re-train the model with trading off the learned knowledge and new knowledge. Kirkpatrick \textit{et. al} \cite{kirkpatrick2017overcoming} propose Elastic Weights Consolidation (EWC), which uses sequential Bayesian inference to approximate the posterior distribution by taking the learned parameters as prior knowledge. EWC finds the important parameters to the learned tasks according to Fisher Information and mitigates their changes by adding quadratic items in the loss function. Similarly, Zenke \textit{et. al} \cite{zenke2017continual} inequitably penalise the parameters in the objective function. Zenke \textit{et. al} define a set of influential parameters by using the information obtained from the gradients of the model. The idea of using a quadratic form to approximate the posterior function is also used in Incremental Moment Matching (IMM) \cite{lee2017overcoming}. In IMM, there are three transfer techniques: weight-transfer, L2-transfer and drop-transfer to smooth the loss surface between the different tasks. The IMM method approximates the mixture. Recently, the variational inference has drawn attention to solving the continual learning problem \cite{nguyen2017variational}. The core idea of this method is to approximate the intractable true posterior distribution by variational learning. Regularisation approach can continually learn new tasks without saving the trained samples or adding new neuron resources. However, when the number of tasks increases, regulating the model becomes very complex.

\textbf{Memory Replay:} The core idea of memory replay is to interleave the new training data with the previously learned samples. The recent developments in this direction reduce the memory of the old knowledge by leveraging a pseudo-rehearsal technique \cite{robins1995catastrophic}. Instead of explicitly storing the entire training samples, the pseudo-rehearsal technique draws the training samples of the old knowledge from a probabilistic distribution model. Shin \textit{et. al} \cite{shin2017continual} propose an architecture consisting of a deep generative model and a task solver. Similarly, Kamra \textit{et. al} \cite{kamra2017deep} use a variational autoencoder to regenerate the previously trained samples. The performance of memory replay approaches is high \cite{shin2017continual,kamra2017deep}. However, it is a memory consuming approach. Furthermore, the computational resources required to train a generative model can also be very high.

\textbf{Dynamic Networks:} Dynamic Networks allocate new neuron resources to learn new tasks. For example, ensemble methods build a network for each task. As a result, the number of models grows linearly with respect to the number of tasks \cite{wozniak2014survey,polikar2001learn++,Dai}. This is not always a desirable solution because of its high resource demand and complexity \cite{kemker2017measuring}. One of the key issues in the dynamic methods is that whenever there is a new task, new neuron resources will be created without considering the possibility of generating redundant resources. In \cite{yoon2017lifelong}, the exponential parameter and resource increases are avoided by selecting part of the existing neurons for training new tasks. However, during the test process, the model has to be aware of which test task is targeted to choose the appropriate parameters to perform the desired task \cite{rusu2016progressive}. In the most dynamic methods \cite{rusu2016progressive,yoon2017lifelong}, the model will not forget the learned knowledge because of the trained parameters are fixed. However, the major issues in dynamic networks are how to prevent the parameters growing exponentially and how to decide which parameters should be used at the testing stage.

The works mentioned above focus on supervised-learning. There are also other exiting works that concentrate on unsupervised learning methods. For example, Bianchi \textit{et. al}\cite{bianchi2019energy} mix a supervised Convolutional Neural Network (CNN) with a bio-inspired unsupervised learning component. Similarly, Mu{\~n}oz-Mart{\'\i}n \textit{et. al} \cite{munoz2019unsupervised} present a novel architecture that combines CNN with unsupervised learning by spike-timing-dependent plasticity (STDP) to overcome the forgetting problem in learning models.

In this paper, we propose a Continual Bayesian Learning Network (CBLN) to address the forgetting problem and to allow the model to adapt to new distributions and learn new tasks. The CBLN trains an entirely new model for each task and merges them into a master model. The master model finds the similarities and distinctions among these sub-models. For the similarities, the master model merges them and produces a general representation. For the distinctive parameters, the master model does not merge them and retains them. CBLN is based on Bayesian Neural Networks (BNNs) \cite{blundell2015weight}, see Figure \ref{fig:network_archi}. Based on BNNs, we assume that the weights in our BNN model have a Gaussian distribution and the covariance matrix is diagonal. The distribution of the weights in different tasks are independent of each other. Based on this assumption, we can assume that the combined posterior distribution of all the training tasks is a mixture of Gaussian distributions. We then use an Expectation-Maximisation (EM) \cite{moon1996expectation} algorithm to approximate the posterior mixture distributions and remove the components that are redundant or less significant. The final distribution of the weights can be a Gaussian mixture distribution with an arbitrary number of components. At the test stage, we produce an epistemic uncertainty \cite{kendall2017uncertainties} measure for each set of components. The set which has minimal uncertainty will be used to give the final prediction. 

 In general, there are two main challenges while solving the continual learning problem with dynamic methods: 1). How to prevent the exponential increase in the number of parameters; 2). How to choose the parameters corresponding to the test task. Firstly, we address the issue of the exponential increase in the number of parameters by using variational learning and a clustering algorithm. These allow us to decrease the number of parameters significantly. Secondly, we address the issue of choosing the parameters by using an uncertainty criterion. Instead of averaging the prediction of different models corresponding to different tasks, or explicitly indicating the task identified in the test state, the proposed model can evaluate the uncertainty of the test points.

\section{Continual Bayesian Learning Networks (CBLN)}
\subsection{Training Process}
The training process in CBLN is similar to BNNs. At the beginning of the training for each task, we initialise all the training parameters and train the model. However, at the end of the training for each task, we store the solution for the current task. We used the loss function shown in Equation (\ref{eq:cbln_loss}):
\begin{equation}
\label{eq:cbln_loss}
\begin{split}
\mathbb{L}(D,\theta) = \mathbb{E}_{q(w|\theta)}[\log P(D|\textbf{w}^{(i)}] &- \\
\frac{\lambda}{2} * D_{KL}&[q(\textbf{w}^{(i)}|\theta)||\log P(\textbf{w}^{(i)}]
\end{split}
\end{equation}
Where $\theta$ refers to the training parameters, $\textbf{w}^{(i)}$ is the $i_{th}$ Monte Carlo sample \cite{hastings1970monte} drawn from the variational posterior $q(\textbf{w}^{(i)}|\theta)$, $D$ is the training data, $\lambda$ is a hyper-parameter to regular the training of models. The common used $\lambda$ is 1. We attempt to obtain weight parameters that have a similar Gaussian distribution, which is close to the prior knowledge. After training $K$ tasks, we can obtain $K$ sets of parameters that construct the posterior mixture Gaussian distribution in which each component is associated with a different task. 

\subsection{Merging Process}
\label{sec:merging_process}
The merging process in this method reduces the components in the posterior mixture distribution. Taking one mixture Gaussian distribution as an example, we approximate the posterior mixture distribution with an arbitrary number of Gaussian distributions, see Equation (\ref{eq:approx_mixture}), where $K$ is the number of tasks, $n$ is the number of components in the final posterior mixture distribution, $q_{1:K}$ is the posterior mixture distribution with the component $q_k$ associated with $k_{th}$ task, $\alpha$ and $\beta$ are the weight parameters where $\alpha = \frac{1}{K},\beta = \frac{1}{n}$. In the extreme case, when $n=1$, this process can be interpreted as a special case of IMM which merges several models into a single one. When $n=K$, this process can be interpreted as a special case of ensemble methods since there are $K$ set of parameters without being merged. 
\begin{equation}
\label{eq:approx_mixture}
    q_{1:K} = \sum_j^K \alpha q_j \approx q_{1:n} = \sum_j^n \beta q_j \text{   where    }  n<=K
\end{equation}
To obtain the final posterior distribution $q_{1:n}$ and restrict the sudden increase in the number of parameters, we approximate the $q_{1:K}$ by using a Gaussian Mixture Model (GMM) \cite{reynolds2015gaussian} with EM algorithm to get $q^*_{1:K}$. We then remove the redundant distributions in $q^*_{1:K}$. 

The EM algorithm contains an Estimation step (E-step) and a Maximisation step (M-step). For each weight, we first sample $N$ data points $x_{[1:N]}$ from the posterior mixture distribution and initialise a GMM model with $K$ components. Then, the E-step estimates the probability of each data point generated from each $K$ random Gaussian distribution, see Equation (\ref{eq:em_estep}). For the $i_{th}$ data point $x_i$, we assume that it is generated from the $m_{th}$ Gaussian distribution and calculate the probability $\pi_m$. We can obtain a matrix of membership weights after applying Equation (\ref{eq:em_estep}) to each data point and determine the mixture of Gaussian distributions. The M-step modifies the parameters of these $K$ random Gaussian distributions by maximising the likelihood according to the weights generated from the first step; see Equation (\ref{eq:em_mstep}). 
\begin{equation}
\label{eq:em_estep}
    \pi_m(x_i) = \frac{\alpha_m \mathbb{N}(x_i|\mu_m,\Sigma_m)}{\sum_{j=1}^K\alpha_j\mathbb{N}(x_i|\mu_j,\Sigma_j)}
\end{equation}

\begin{equation}
\label{eq:em_mstep}
\begin{split}
        \alpha_j &= \frac{1}{N} \sum_{i=1}^N \pi_m(x_i), \\    
        \mu_j &= \frac{\sum_{i=1}^N\pi_m(x_i)x_i}{\sum_{i=1}^N\pi_m(x_i)},  \\   
    \Sigma &= \frac{\sum_{i=1}^N\pi_m(x_i)(x_i-\mu_j)(x_i-\mu_j)^T}{\sum_{i=1}^N\pi_m(x_i)}
\end{split}
 \end{equation} 
 
 After the algorithm is converged, we can obtain an approximated posterior mixture distribution $q^*_{1:K} = \sum_j^K \alpha^* q^*_j $, where $\sum_j^K \alpha_j^* = 1$. We then remove $q^*_j $, if $\alpha_j^*$ is smaller than a threshold which is set to $t=\frac{1}{2K}$.  These distributions can be regarded as redundant components which overfit the model. Since the EM algorithm clusters similar data points into one cluster, we can merge the distributions if they are similar to each other and get the final posterior mixture distribution $q_{1:n}$. We use the trained GMM to cluster the mean value of each component in $q_{1:K}$. If the mean values of two distributions are in the same cluster, these two distributions are merged into a single Gaussian distribution. The training and merging processes are recursive. In other words, the model saves information about learned mixture distributions after learning several tasks. When there is a new task, the model learns the new tasks and then merges the new distributions with existing Gaussian mixture distributions.
 \subsection{Testing Process}
After the training and merging processes, we obtain sets of parameters to construct the mixture posterior distribution with several components. Since the CBLN contains the information from different learned tasks, we need to identify which information is suitable to give a prediction in the test state. We obtain several Monte Carlo samples of the weights drawn from the variational posterior to determine the uncertainty. For this purpose, we calculate the variance of the predictive scores. The set of parameters which has minimal uncertainty is chosen to give the final prediction. We use the epistemic uncertainty in our calculation. There are also other uncertainty measurements such as computing the entropy of the predictive scores \cite{renyi1961measures} or Model Uncertainty as Measured by Mutual Information (MUMMI) \cite{rawat2017adversarial}, see Equation (\ref{eq:mummi}). The trade-off between these uncertainty measures is discussed in Section \ref{sec:dis}.

\begin{equation}
\label{eq:mummi}
\begin{split}
    \textbf{Entropy} &= \mathbb{H}[y^*|x,D]\\
    \textbf{MUMMI} &= \mathbb{H}[y^*|x,D] + \mathbb{E}_{q(w^{(i)}|\theta)}[\mathbb{H}[y^*|x,w^{(i)}]]
\end{split}
\end{equation}

Where $y^*$ is the predicted distribution, $q(\textbf{w}^{(i)}|\theta)$ is the variational posterior distribution, and $x$ is the test input.

\section{Experiments}
We evaluated our method on the MNIST \cite{lecun2010mnist} image datasets and the UCR Two Patterns time-series dataset \cite{UCRArchive}. MNIST is a handwritten dataset consist of 10 digits. UCR is an archive that contains batches of time-series datasets. The MNIST and Two Patterns contain 60000 and 1000 training samples, 10000 and 4000 test samples, 10 and 4 classes respectively.

In our experiments, we do not re-access the samples after the first training but let the model know that it needs to train for a new task. However, the difference in our method compared with the existing works is that we do not tell the model which task is being tried. Furthermore, the output nodes refer to the appropriate number of classes that the task is trained for. The overlap between the output classes, which are trained at different times, are also taken into consideration. This means that at the time of the training for each task, we do not know which other tasks the new samples could also be associated with. The settings in our experiments are similar to \cite{lee2017overcoming}, which is more strict than other settings in the existing works. For example, in contrast to our experiments, the other existing experiments are allowed to re-access the training samples \cite{shin2017continual}, tell the model which task is the test data comes from \cite{rusu2016progressive}, or use different classifiers for different tasks \cite{nguyen2017variational}.

In CBLN, we randomly choose 200 test data from the test task and draw 200 Monte Carlo samples from the posterior distribution and measured the uncertainty to decide which parameters should be used in the model for each particular task. 

We compare our model with state-of-the-art methods including Neural Networks (NN), Incremental Moment Matching (IMM) \cite{lee2017overcoming} and Synaptic Intelligence (SI) \cite{zenke2017continual}. In the IMM model, we perform all the IMM algorithms combining with all the transfer techniques mentioned in \cite{lee2017overcoming}. We also search for the best hyper-parameters and choose the best accuracy according to \cite{lee2017overcoming}. In the SI model, we search the best hyper-parameters as well. For the SI, Multiple-Head (MH) approach is used in the original paper. The MH approach is used to divide the output layer into several sections. For different tasks, each section will be activated in which the overlap between different classes in different tasks is also avoided. MH approach requires the model to be told about the test tasks. We perform our evaluation based on the SI approach with and without using the MH approach. In CBLN, we search for the best model that can distinguish the test data. Since CBLN is based on BNN, we use the BNN as a baseline for comparison.

\subsection{Split MNIST}
The first experiment is based on the split MNIST dataset. This experiment is to evaluate the ability of the model to learn new tasks continually. In this experiment, we split the MNIST dataset into several sub-sets; e.g. when the number of tasks is one, the networks are trained on the original MNIST; when the number of tasks is two, the network is trained on the digits 0 to 4, 5 to 9 sequentially. The rest of the process follows the same manner. As the number of tasks increasing, we keep splitting the MNIST into multiple sub-sets. To implement the other methods, we follow the optimal architecture described in the original papers. In IMM, we use two hidden layers with 800 neurons each. In SI and NN, we use two hidden layers with 250 neurons each. We use a BNN, which contains two hidden layers with 25 neurons in each layer. The total number of parameters in the BNN is 41070 (the number of parameters in BNNs is doubled). In CBLN, we use two hidden layers with only ten neurons each. To evaluate the performance, we compute the average of test accuracy on all the tasks.
\begin{figure}
    \centering
    
  \subfloat[Test Accuracy\label{fig:res_split_mnist_acc}]{%
       \includegraphics[width=\linewidth]{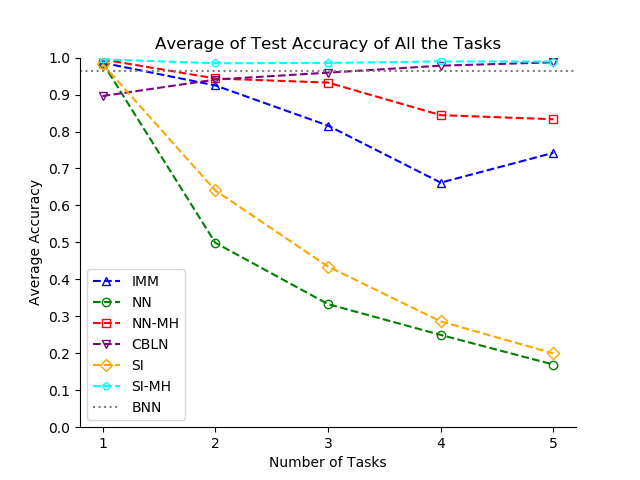}}
      
     \subfloat[Number of Parameters\label{fig:res_split_mnist_para}]{%
       \includegraphics[width=\linewidth]{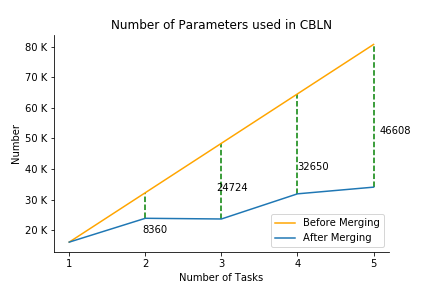}}
       
         \subfloat[Uncertainty\label{fig:res_split_mnist_uncertain}]{%
       \includegraphics[width=\linewidth]{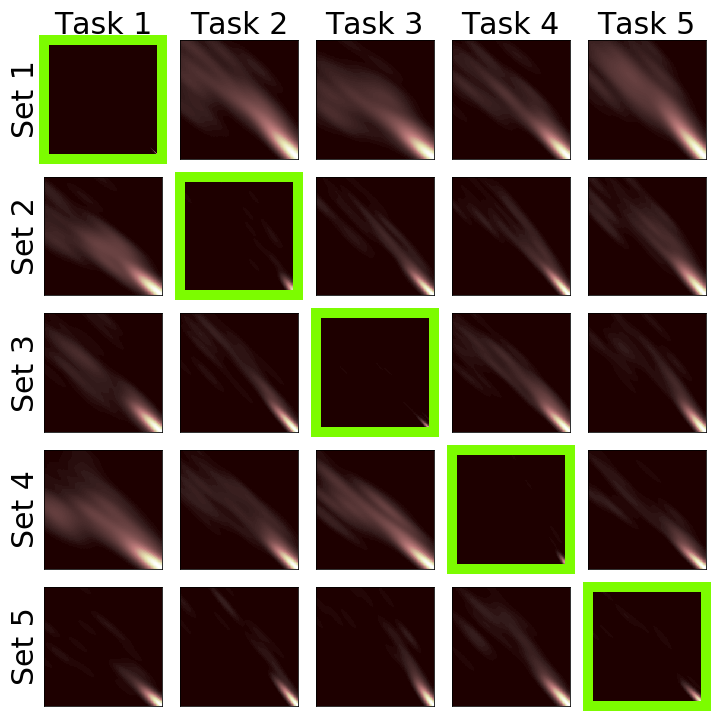}}
       
    \caption{Split Mnist Experiment. (a) Average of test accuracy of all the tasks. (b) The number of parameters in CBLN before and after the merging process. (c) Uncertainty of the model on the test tasks when the number of tasks is set to 5.}
    \label{fig:res_split_mnist_analyse}
\end{figure}
As shown in Figure \ref{fig:res_split_mnist_acc}, the average test accuracy of all the tasks in CBLN keeps increasing, while the performance of other methods decreases over time. As long as we divide the MNIST model training into several simpler tasks, the performance of CBLN keeps increasing since CBLN can learn the new tasks without forgetting the previously learned ones. The accuracy after training five tasks sequentially reaches the performance of SI with the MH approach. Shown in Figure \ref{fig:res_split_mnist_acc}, the method using the MH approach avoids the interference between the tasks with different classes at the output layer (i.e. by interference we mean the situation in which learning a new task causes changing the parameters in a way that the model forgets the previously learned ones). However, we need to tell the model which task the test data refers to in both training and test processes. The grey line in Figure \ref{fig:res_split_mnist_acc} represents the accuracy of training a BNN with the original MNIST. The performance of CBLN which continually learns five different tasks outperforms the BNN.

The parameters used in CBLN are less than the BNN. Figure \ref{fig:res_split_mnist_para} illustrates the number of parameters used in CBLN. The orange line shows the number of parameters before the merging process. The blue line shows the number of parameters after the merging process, and the green lines illustrate the number of merged parameters. The number of parameters used while training five tasks is 35094, which is significantly lower than the parameters used in other state-of-the-art methods. The CBLN only doubles the number of parameters during the experiment (which is 16140 at the beginning). The more tasks are trained, the more parameters are merged because CBLN finds the similarity among the solutions for all the tasks and merges them. 

Figure \ref{fig:res_split_mnist_uncertain} illustrates the uncertainty measure in the test process when the number of tasks is five. In each block, the x-axis shows the prediction score for that particular task; the y-axis shows the variance. If the density of highlighted points is close to the lower right corner, the model has low uncertainty and high prediction score. The blocks shown in the diagonal line are the results with the lowest level of uncertainty for each particular task. 
\subsection{Permuted MNIST}


The second experiment is based on the permuted MNIST to evaluate the ability of the model to learn new tasks incrementally. This experiment is different from the split MNIST experiment since the number of classes in each task is always 10. We follow the same setting in the previous work done by Kirkpatrick \textit{et. al}, Lee \textit{et. al} in \cite{kirkpatrick2017overcoming,lee2017overcoming}. The first task is based on the original MNIST. In the rest of the tasks, we shuffle all the pixels in the images with different random seeds. Therefore, each task requires a different solution. However, the difficulty level of all the tasks is similar. In this experiment, CBLN contains two hidden layers, with each having 50 neurons. 

\begin{figure}
    \centering
    
         \subfloat[Test Accuracy\label{fig:res_permute_1_acc}]{%
       \includegraphics[width=\linewidth]{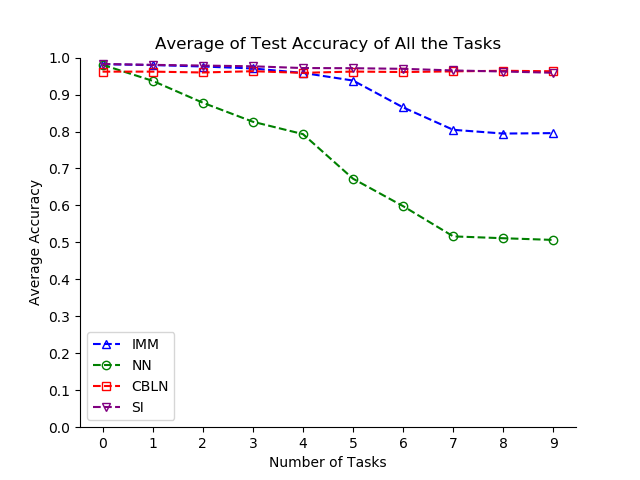}}
       
         \subfloat[Number of Parameters\label{fig:res_permute_1_params}]{%
       \includegraphics[width=\linewidth]{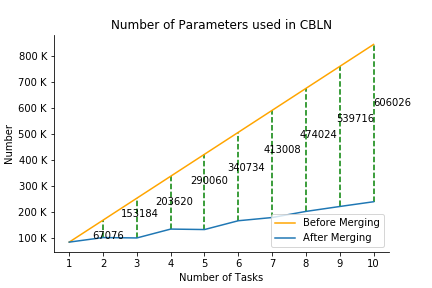}}
       
    \caption{Permuted MNIST experiment. }
    \label{fig:res_permute}
\end{figure}

\begin{figure}
    \centering
         \subfloat[Test Accuracy\label{fig:res_ucr_acc}]{%
       \includegraphics[width=\linewidth]{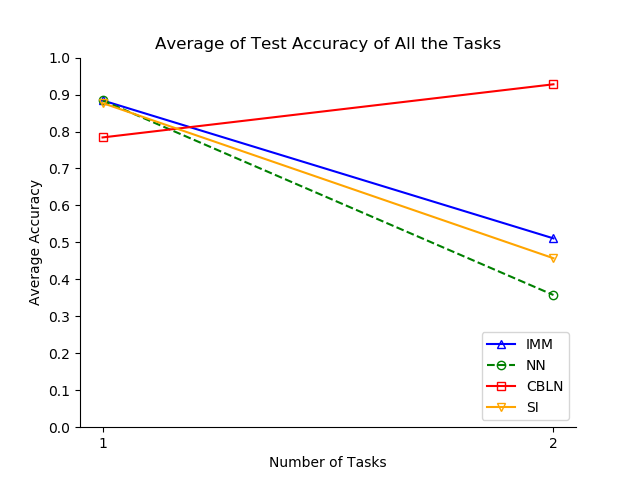}}
       
         \subfloat[Number of Parameters\label{fig:res_ucr_params}]{%
       \includegraphics[width=\linewidth]{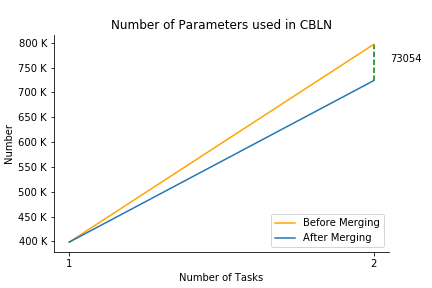}}
       
         \subfloat[Uncertainty Changes\label{fig:res_ucr_uncertain}]{%
       \includegraphics[width=\linewidth]{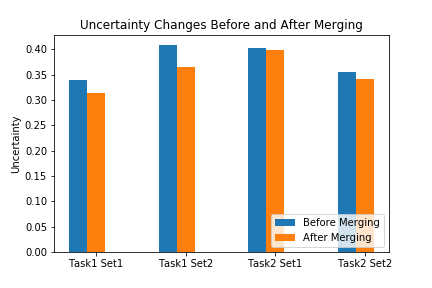}}
 \caption{Time Series Experiment. (a) Average of test accuracy of all the tasks. (b) The number of parameters in CBLN before and after the merging process. (c) Uncertainty changes before and after the merging process.}
     \label{fig:res_ucr}
\end{figure}


\subsection{Time-Series data}
In the last experiment, we use the Two-Patterns dataset from UCR time-series archive. In this experiment, CBLN uses two hidden layers, each containing 200 neurons. The other methods with two hidden layers, each containing 800 neurons with Dropout layers. While training the CBLN model with the entire Two-Patterns dataset, the best accuracy is around 0.8. If we split the dataset into two parts, the accuracy is above 0.9. The accuracy of CBLN outperforms other methods by continually learning Two-Patterns dataset divided into smaller tasks rather than learning it as an entire model.

\begin{figure}
    \centering
         \subfloat[Task 1\label{fig:an_s1}]{%
       \includegraphics[width=\linewidth]{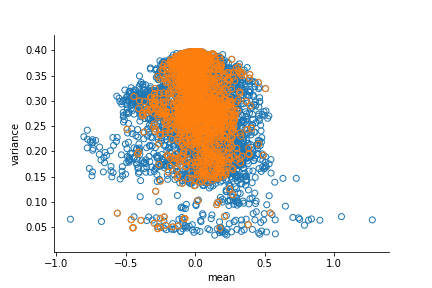}}
       
         \subfloat[Task 2\label{fig:an_s2}]{%
       \includegraphics[width=\linewidth]{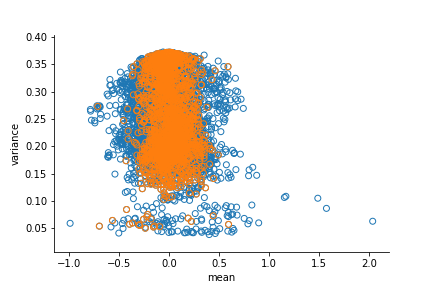}}
       
         \subfloat[Task 1 and Task 2\label{fig:an_dm12}]{%
       \includegraphics[width=\linewidth]{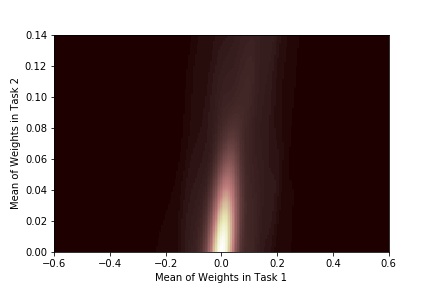}}
    \caption{From split MNIST experiments. Orange points represent the merged weights. (a) Scatter plot of the weights in Task 1. (b) Scatter plot of the weights in Task 2. (c) Density of the merged weights for task 1 and 2.}
    \label{fig:an_merged}
\end{figure}

\begin{figure}
    \centering
         \subfloat[Absolute Difference\label{fig:res_split_mnist_diff_acc}]{%
       \includegraphics[width=\linewidth]{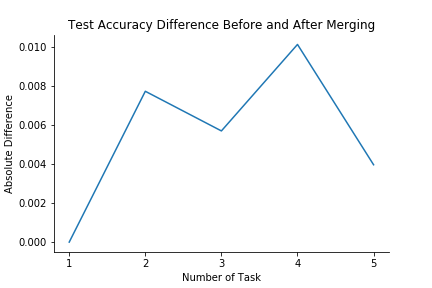}}
       
         \subfloat[Time Cost\label{fig:res_split_mnist_time_cost}]{%
       \includegraphics[width=\linewidth]{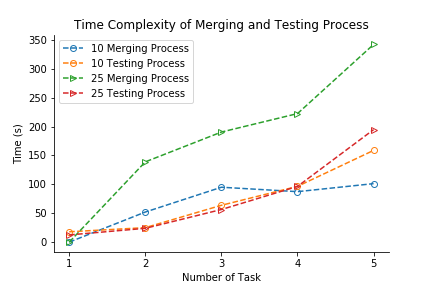}}
       
         \subfloat[Variance. Correct: 4/10\label{fig:res_split_mnist_uncertain_10Task}]{%
       \includegraphics[width=0.8\linewidth]{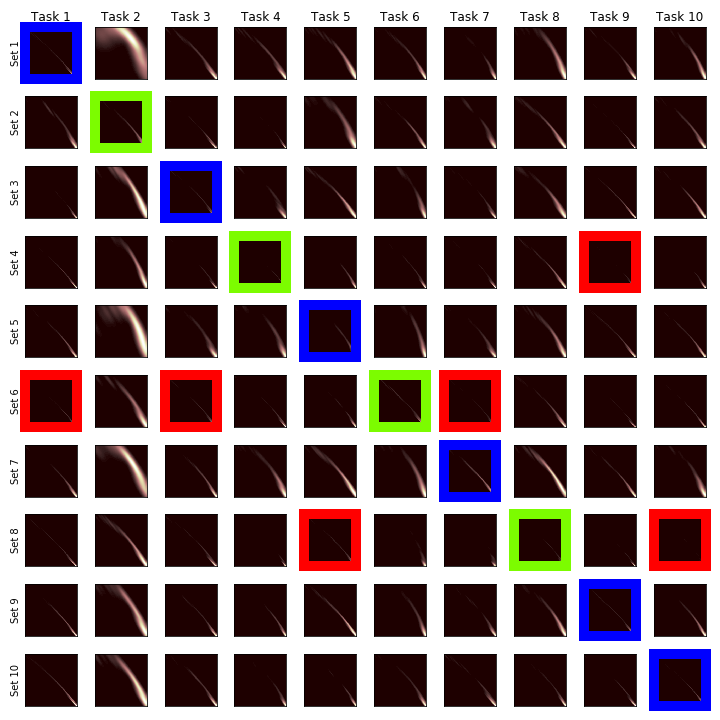}}
    \caption{Based on the split MNIST experiment. (a) Absolute difference between the test accuracy before and after the merging process. The maximum value in the y-axis is 0.01. (b) The running time of merging and testing process. 10 and 25 are the number of neurons in each layer. (c) Uncertainty of test task when the number of tasks is ten. Each task contains only one class.}
    \label{fig:res_split_mnist}
\end{figure}

\section{Discussion}
\label{sec:dis}
\textbf{Merged weights}: We start the discussion with analysing how the weights are merged. We visualise the weights in the Split MNIST experiment that was carried out with two tasks. Shown in Figure \ref{fig:an_merged}, the orange points represent the merged weights. In Figure \ref{fig:an_s1},\ref{fig:an_s2}, the x-axis shows the mean of weights; the y-axis shows the variance of the weights. Figure \ref{fig:an_dm12} shows the density of the merged parameters. If the mean of the weight distribution is closer to 0, the weight has a larger chance to be merged because our prior knowledge is a Gaussian distribution with a mean of 0. For the weights which the mean values are higher, they have less chance to be merged because these weights can be regarded as to have larger contributions to finding the solution for the training tasks. For each task, the solution could be different. Hence these weights are the distinctions among different tasks. 

\textbf{Sequential Updating}: The settings of experiments above are the same as described in Lee \textit{et. al}\cite{lee2017overcoming}. These experiments inform the model about the number of tasks is it will perform. The experiments also merge all into a single model. This strategy can be viewed as training the tasks in parallel. We perform further evaluations on giving the tasks in sequential order. In this case, the model learns a new task and merges it with existing knowledge at each time slot. We train the 5 tasks of split MNIST and Permuted MNIST sequentially and compare to the parallel learning manner. The results are shown in Table \ref{tab:parallel_sequential}.
\begin{table}[]
    \centering
    \caption{Comparison the parallel learning with sequential learning.}
    \begin{tabular}{|c|c|c|}
    \hline
         &  Parallel & Sequential \\
    \hline
    Split MNIST     & 97.14\% & 97.69\% \\
    \hline
    Permuted MNIST & 96.02\% & 95.86\% \\ 
    \hline
    \end{tabular}
    \label{tab:parallel_sequential}
\end{table}

\textbf{Ablation study}: Inspired by \cite{kemker2017measuring}, we have evaluated our model with and without the merging process. To evaluate the performance decreases after merging the models, we calculate the absolute difference of test accuracy before and after the merging process. Shown in Figure \ref{fig:res_split_mnist_diff_acc}, the absolute difference is almost 0. Therefore, all the similar parameters have been merged perfectly, and the distinct parameters are maintained very well. To evaluate the uncertainty changes before and after the merging process, we track the uncertainty changes in the time-series experiment. Shown in Figure \ref{fig:res_ucr_uncertain}, the uncertainties are decreased after the merging process, but it can still help the model to choose the correct parameters to predict the test data. Furthermore, the merging process significantly decreases the number of parameters needed to learn a mode for multiple tasks as shown in Figure \ref{fig:res_split_mnist_para},\ref{fig:res_permute_1_params},
\ref{fig:res_ucr_params}. The merging process can significantly prevent the exponential increase in the number of parameters required to learn the model without degrading the performance. The main advantages of CBLN compared with other dynamic methods are choosing suitable parameters to classify the test samples and preventing an exponential increase in the number of parameters. We describe  Progressive Neural Networks (PNNs) as an example. In PNNs, if the initial number of parameters in the model is $N$, after learning $K$ tasks, the number of parameters in the model increases to $N*K$. The number of parameters needed are shown in orange lines in Figure \ref{fig:res_split_mnist_para},\ref{fig:res_permute_1_params},
\ref{fig:res_ucr_params}. Furthermore, the model needs to be informed about which task is currently given in advance at the test stage.

\textbf{Complexity}: We then evaluate the time complexity of the merging and testing process with respect to the initial neuron resources. We ran the experiments on a Macbook Pro (2015) with 2.7 GHz Intel Core i5. Shown in Figure \ref{fig:res_split_mnist_time_cost}, where 10 represents the CBLN contains two hidden layers with ten neurons each, 25 represents the CBLN contains two hidden layers with 25 neurons each. CBLN is time-consuming during the test state, especially when the number of trained tasks grows. To produce the uncertainty measure, the computational complexity of CBLN is $O(n^2)$ while the BNNs are $O(n)$ for each test data. We assume the model does not know in advance which task the test data is associated with. This means that the model needs to identify and chooses the correct solution for each test task. This is a key advantage of CBLN compared to other existing methods that assume the model knows in advance which test task is being performed; e.g. \cite{rusu2016progressive,nguyen2017variational,zenke2017continual}. The test stage could be the same as a conventional neural network if we informed the model which task is being tested. However, in real-world applications, this information is not available to the model in advance. The CBLN uses the uncertainty measure to choose the appropriate learned solution for each particular task. The number of tasks does not have much effect on the merging process. The main effect on the merging process is the number of parameters of the model at the initialisation. According to our experiments, we can initialise CBLN with a much smaller number of parameters to solve a complex task as long as it can solve it as a set of simpler tasks. Furthermore, CBLN does not need to evaluate the importance of parameters by measures such as computing Fisher Information (second derivative of loss function) \cite{lee2017overcoming,kirkpatrick2017overcoming} which are computationally expensive and intractable in large models. In summary, the testing process of CBLN is increasing with respect to the number of tasks having been learned; the merging process of CBLN is increasing with respect to the initial neuron resources.

\textbf{Uncertainty measure}: In this section, we discuss the epistemic uncertainty measure that is computed by the model given test data. The CBLN uses epistemic uncertainty measure to identify the current task form the distribution of training data. We evaluate the variance, entropy and MUMMI in different experiments. To see which measurement of uncertainty is suitable to be used in CBLN for choosing the learned solution, we run each experiment for ten times and calculate the average selection rate. Shown in Table \ref{tab:un_eval}, in the permuted MNIST experiments, although the number of tasks is increasing, the model can choose the correct solutions. In the split MNIST experiment, the rate of uncertainty decreases, if the number of tasks increases. In other words, the model cannot distinguish the tasks that the test data is associated with when the number of classes in each task decrease. We analyse this as a Rare Class Scenario in Epistemic Uncertainty (RCSEU). RCSEU means that when the number of classes in each task is very small, the model will overfit the training data quickly and will become over-confident with the result of classifying the test data.

To illustrate the RCSEU, we visualise the uncertainty information in the split MNIST experiment, when the number of tasks is ten. In Figure \ref{fig:res_split_mnist_uncertain_10Task}, the blue blocks are the correct solution (in the diagonal line), the green blocks represent that the model identify the test data correctly and the red blocks represent that the model identifies the test data incorrectly and the black blocks represent very small uncertainty.

\begin{table}[]
    \centering
    \caption{The average rate of correct selection. }
    \begin{tabular}{|c|c|c|c|c|c|c|}
    \hline  
    Experiment  &  \multicolumn{5}{|c|}{Split MNIST} & Permuted \\\cline{1-7}
                
    Number of Tasks &  2  &3   & 4   &5  & 10 & 10           \\
    \hline  
    Variance    & 1.0   &1.0  &0.95  &0.866 & 0.3   & 1.0              \\
    Entropy     & 1.0   &0.8  &0.7   &0.736  & 0.29   & 1.0            \\
    MUMMI       & 1.0   &0.87 &0.925  &0.894 & 0.32  & 1.0             \\
    \hline
    \end{tabular}
    \label{tab:un_eval}
\end{table}

\section{Conclusions}
This paper proposes the Continual Bayesian Learning Networks (CBLN) to solve the forgetting problem in continual learning scenarios. CBLN is based on Bayesian Neural Networks (BNNs). Different from BNNs, the weights in the CBLN are mixture Gaussian distributions with an arbitrary number of distributions. The CBLN can solve a complex task by dividing it into several simpler tasks and learning each of them sequentially. Since CBLN uses mixture Gaussian distribution models in its network, the number of additionally required parameters decreases as the number of tasks increases. The CBLN identifies which solution should be used for which test data by using an uncertainty measure. More importantly, our proposed model can overcome the forgetting problem in learning models without requiring to re-access previously seen training samples. We have evaluated our method based on MNIST image and UCR time-series datasets and have compared the results with the state-of-the-art models. In the split MNIST experiment, our method outperforms the Incremental Moment Matching (IMM) model by 25\%, and the Synaptic Intelligence (SI) model by 80\%. In the permuted MNIST experiment, our method outperforms IMM by 16\% and achieves the same accuracy as the SI model. In the time-Series experiment, our method outperforms IMM by 40\% and the SI model by 47\%. The future work will focus on developing solutions to let the model determine when it needs to train for a new task given a series of samples. This will be achieved by analysing the drifts and changes in the distribution of the training data. The work will also focus on developing methods to group the neurons during the merging process to construct regional functional areas in the network specific to a set of similar tasks. This will allow us to reduce the complexity of the network and create more scalable and generalisable models. 

\section*{Acknowledgment}
This work is partially supported by the Care Research and Technology Centre at the UK Dementia Research Institute (UK DRI).

\ifCLASSOPTIONcaptionsoff
  \newpage
\fi



\bibliographystyle{IEEEtran}
\bibliography{ref}

\begin{thebibliography}{10}
\providecommand{\url}[1]{#1}
\csname url@samestyle\endcsname
\providecommand{\newblock}{\relax}
\providecommand{\bibinfo}[2]{#2}
\providecommand{\BIBentrySTDinterwordspacing}{\spaceskip=0pt\relax}
\providecommand{\BIBentryALTinterwordstretchfactor}{4}
\providecommand{\BIBentryALTinterwordspacing}{\spaceskip=\fontdimen2\font plus
\BIBentryALTinterwordstretchfactor\fontdimen3\font minus
  \fontdimen4\font\relax}
\providecommand{\BIBforeignlanguage}[2]{{%
\expandafter\ifx\csname l@#1\endcsname\relax
\typeout{** WARNING: IEEEtran.bst: No hyphenation pattern has been}%
\typeout{** loaded for the language `#1'. Using the pattern for}%
\typeout{** the default language instead.}%
\else
\language=\csname l@#1\endcsname
\fi
#2}}
\providecommand{\BIBdecl}{\relax}
\BIBdecl

\bibitem{deng2009imagenet}
J.~Deng, W.~Dong, R.~Socher, L.-J. Li, K.~Li, and L.~Fei-Fei, ``Imagenet: A
  large-scale hierarchical image database,'' in \emph{2009 IEEE conference on
  computer vision and pattern recognition}.\hskip 1em plus 0.5em minus
  0.4em\relax Ieee, 2009, pp. 248--255, doi: 10.1109/CVPR.2009.5206848.

\bibitem{he2016deep}
K.~He, X.~Zhang, S.~Ren, and J.~Sun, ``Deep residual learning for image
  recognition,'' in \emph{Proceedings of the IEEE conference on computer vision
  and pattern recognition}, 2016, pp. 770--778, doi: 10.1109/CVPR.2016.90.

\bibitem{shalev2011online}
S.~Shalev-Shwartz \emph{et~al.}, ``Online learning and online convex
  optimization,'' \emph{Foundations and trends in Machine Learning}, vol.~4,
  no.~2, pp. 107--194, 2011, doi: 10.1561/2200000018.

\bibitem{de2019continual}
M.~De~Lange, R.~Aljundi, M.~Masana, S.~Parisot, X.~Jia, A.~Leonardis,
  G.~Slabaugh, and T.~Tuytelaars, ``Continual learning: A comparative study on
  how to defy forgetting in classification tasks,'' \emph{arXiv preprint
  arXiv:1909.08383}, 2019.

\bibitem{mermillod2013stability}
M.~Mermillod, A.~Bugaiska, and P.~Bonin, ``The stability-plasticity dilemma:
  Investigating the continuum from catastrophic forgetting to age-limited
  learning effects,'' \emph{Frontiers in psychology}, vol.~4, p. 504, 2013,
  doi: 10.3389/fpsyg.2013.00504.

\bibitem{ditzler2015learning}
G.~Ditzler, M.~Roveri, C.~Alippi, and R.~Polikar, ``Learning in nonstationary
  environments: A survey,'' \emph{IEEE Computational Intelligence Magazine},
  vol.~10, no.~4, pp. 12--25, 2015, doi: 10.1109/MCI.2015.2471196.

\bibitem{mccloskey1989catastrophic}
M.~McCloskey and N.~J. Cohen, ``Catastrophic interference in connectionist
  networks: The sequential learning problem,'' in \emph{Psychology of learning
  and motivation}.\hskip 1em plus 0.5em minus 0.4em\relax Elsevier, 1989,
  vol.~24, pp. 109--165, doi: "10.1016/S0079-7421(08)60536-8".

\bibitem{goodfellow2013empirical}
I.~J. Goodfellow, M.~Mirza, D.~Xiao, A.~Courville, and Y.~Bengio, ``An
  empirical investigation of catastrophic forgetting in gradient-based neural
  networks,'' \emph{arXiv preprint arXiv:1312.6211}, 2013.

\bibitem{mcclelland1995there}
J.~L. McClelland, B.~L. McNaughton, and R.~C. O'Reilly, ``Why there are
  complementary learning systems in the hippocampus and neocortex: insights
  from the successes and failures of connectionist models of learning and
  memory.'' \emph{Psychological review}, vol. 102, no.~3, p. 419, 1995, doi:
  10.1037/0033-295X.102.3.419.

\bibitem{barnett2002and}
S.~M. Barnett and S.~J. Ceci, ``When and where do we apply what we learn?: A
  taxonomy for far transfer.'' \emph{Psychological bulletin}, vol. 128, no.~4,
  p. 612, 2002, doi: 10.1037/0033-2909.128.4.612.

\bibitem{parisi2019continual}
G.~I. Parisi, R.~Kemker, J.~L. Part, C.~Kanan, and S.~Wermter, ``Continual
  lifelong learning with neural networks: A review,'' \emph{Neural Networks},
  2019, doi: 10.1016/j.neunet.2019.01.012.

\bibitem{kirkpatrick2017overcoming}
J.~Kirkpatrick, R.~Pascanu, N.~Rabinowitz, J.~Veness, G.~Desjardins, A.~A.
  Rusu, K.~Milan, J.~Quan, T.~Ramalho, A.~Grabska-Barwinska \emph{et~al.},
  ``Overcoming catastrophic forgetting in neural networks,'' \emph{Proceedings
  of the national academy of sciences}, vol. 114, no.~13, pp. 3521--3526, 2017,
  doi: 10.1073/pnas.1611835114.

\bibitem{zenke2017continual}
F.~Zenke, B.~Poole, and S.~Ganguli, ``Continual learning through synaptic
  intelligence,'' in \emph{Proceedings of the 34th International Conference on
  Machine Learning-Volume 70}.\hskip 1em plus 0.5em minus 0.4em\relax JMLR.
  org, 2017, pp. 3987--3995.

\bibitem{lee2017overcoming}
S.-W. Lee, J.-H. Kim, J.~Jun, J.-W. Ha, and B.-T. Zhang, ``Overcoming
  catastrophic forgetting by incremental moment matching,'' in \emph{Advances
  in Neural Information Processing Systems}, 2017, pp. 4652--4662.

\bibitem{nguyen2017variational}
C.~V. Nguyen, Y.~Li, T.~D. Bui, and R.~E. Turner, ``Variational continual
  learning,'' \emph{arXiv preprint arXiv:1710.10628}, 2017, doi:
  10.17863/CAM.35471.

\bibitem{robins1995catastrophic}
A.~Robins, ``Catastrophic forgetting, rehearsal and pseudorehearsal,''
  \emph{Connection Science}, vol.~7, no.~2, pp. 123--146, 1995, doi:
  10.1080/09540099550039318.

\bibitem{shin2017continual}
H.~Shin, J.~K. Lee, J.~Kim, and J.~Kim, ``Continual learning with deep
  generative replay,'' in \emph{Advances in Neural Information Processing
  Systems}, 2017, pp. 2990--2999.

\bibitem{kamra2017deep}
N.~Kamra, U.~Gupta, and Y.~Liu, ``Deep generative dual memory network for
  continual learning,'' \emph{arXiv preprint arXiv:1710.10368}, 2017.

\bibitem{wozniak2014survey}
M.~Wo{\'z}niak, M.~Gra{\~n}a, and E.~Corchado, ``A survey of multiple
  classifier systems as hybrid systems,'' \emph{Information Fusion}, vol.~16,
  pp. 3--17, 2014, doi: 10.1016/j.inffus.2013.04.006.

\bibitem{polikar2001learn++}
R.~Polikar, L.~Upda, S.~S. Upda, and V.~Honavar, ``Learn++: An incremental
  learning algorithm for supervised neural networks,'' \emph{IEEE transactions
  on systems, man, and cybernetics, part C (applications and reviews)},
  vol.~31, no.~4, pp. 497--508, 2001, doi: 10.1109/5326.983933.

\bibitem{Dai}
W.~Dai, Q.~Yang, G.-R. Xue, and Y.~Yu, ``Boosting for transfer learning,'' in
  \emph{Proceedings of the 24th International Conference on Machine Learning},
  ser. ICML '07.\hskip 1em plus 0.5em minus 0.4em\relax New York, NY, USA: ACM,
  2007, pp. 193--200, doi:10.1145/1273496.1273521.

\bibitem{kemker2017measuring}
R.~Kemker, M.~McClure, A.~Abitino, T.~Hayes, and C.~Kanan, ``Measuring
  catastrophic forgetting in neural networks,'' \emph{arXiv preprint
  arXiv:1708.02072}, 2017.

\bibitem{yoon2017lifelong}
J.~Yoon, E.~Yang, J.~Lee, and S.~J. Hwang, ``Lifelong learning with dynamically
  expandable networks,'' \emph{arXiv preprint arXiv:1708.01547}, 2017.

\bibitem{rusu2016progressive}
A.~A. Rusu, N.~C. Rabinowitz, G.~Desjardins, H.~Soyer, J.~Kirkpatrick,
  K.~Kavukcuoglu, R.~Pascanu, and R.~Hadsell, ``Progressive neural networks,''
  \emph{arXiv preprint arXiv:1606.04671}, 2016.

\bibitem{bianchi2019energy}
S.~Bianchi, I.~Mu{\~n}oz-Martin, G.~Pedretti, O.~Melnic, S.~Ambrogio, and
  D.~Ielmini, ``Energy-efficient continual learning in hybrid
  supervised-unsupervised neural networks with pcm synapses,'' in \emph{2019
  Symposium on VLSI Technology}.\hskip 1em plus 0.5em minus 0.4em\relax IEEE,
  2019, pp. T172--T173, doi: 10.23919/VLSIT.2019.8776559.

\bibitem{munoz2019unsupervised}
I.~Mu{\~n}oz-Mart{\'\i}n, S.~Bianchi, G.~Pedretti, O.~Melnic, S.~Ambrogio, and
  D.~Ielmini, ``Unsupervised learning to overcome catastrophic forgetting in
  neural networks,'' \emph{IEEE Journal on Exploratory Solid-State
  Computational Devices and Circuits}, vol.~5, no.~1, pp. 58--66, 2019, doi:
  10.1109/JXCDC.2019.2911135.

\bibitem{blundell2015weight}
C.~Blundell, J.~Cornebise, K.~Kavukcuoglu, and D.~Wierstra, ``Weight
  uncertainty in neural networks,'' \emph{arXiv preprint arXiv:1505.05424},
  2015.

\bibitem{moon1996expectation}
T.~K. Moon, ``The expectation-maximization algorithm,'' \emph{IEEE Signal
  processing magazine}, vol.~13, no.~6, pp. 47--60, 1996, doi:
  10.1109/79.543975.

\bibitem{kendall2017uncertainties}
A.~Kendall and Y.~Gal, ``What uncertainties do we need in bayesian deep
  learning for computer vision?'' in \emph{Advances in neural information
  processing systems}, 2017, pp. 5574--5584.

\bibitem{hastings1970monte}
W.~K. Hastings, ``Monte carlo sampling methods using markov chains and their
  applications,'' 1970, doi: 10.1093/biomet/57.1.97.

\bibitem{reynolds2015gaussian}
D.~Reynolds, ``Gaussian mixture models,'' \emph{Encyclopedia of biometrics},
  pp. 827--832, 2015, doi: 10.1007/978-0-387-73003-5\_196.

\bibitem{renyi1961measures}
A.~R{\'e}nyi \emph{et~al.}, ``On measures of entropy and information,'' in
  \emph{Proceedings of the Fourth Berkeley Symposium on Mathematical Statistics
  and Probability, Volume 1: Contributions to the Theory of Statistics}.\hskip
  1em plus 0.5em minus 0.4em\relax The Regents of the University of California,
  1961.

\bibitem{rawat2017adversarial}
A.~Rawat, M.~Wistuba, and M.-I. Nicolae, ``Adversarial phenomenon in the eyes
  of bayesian deep learning,'' \emph{arXiv preprint arXiv:1711.08244}, 2017.

\bibitem{lecun2010mnist}
Y.~LeCun, C.~Cortes, and C.~Burges, ``Mnist handwritten digit database,''
  \emph{AT\&T Labs [Online]. Available: http://yann. lecun. com/exdb/mnist},
  vol.~2, 2010.

\bibitem{UCRArchive}
Y.~Chen, E.~Keogh, B.~Hu, N.~Begum, A.~Bagnall, A.~Mueen, and G.~Batista, ``The
  ucr time series classification archive,'' July 2015, doi:
  10.1109/JAS.2019.1911747,.

\end{thebibliography}
\end{document}